\documentclass[conference]{IEEEtran}

\usepackage{algorithm}
\usepackage{algorithmic}
\usepackage{amsmath}
\usepackage{amssymb}
\usepackage{color}
\usepackage{gensymb}
\usepackage{graphicx}
\usepackage[bookmarks=true]{hyperref}
\usepackage{multicol}
\usepackage[numbers]{natbib}
\usepackage{stfloats}
\usepackage{subcaption}
\usepackage{times}
\usepackage{wrapfig}
\usepackage{xspace}

\newcommand{\badgr}{BADGR\xspace} 

\newcommand{\ba}{\mathbf{a}}
\newcommand{\bo}{\mathbf{o}}
\newcommand{\be}{\mathbf{e}}

\newcommand{\loss}{\mathcal{L}}
\newcommand{\dataset}{\mathcal{D}}

\begin{document}

\title{BADGR: An Autonomous Self-Supervised Learning-Based Navigation System}

\author{\authorblockN{Gregory Kahn, Pieter Abbeel, Sergey Levine}
\authorblockA{Berkeley AI Research (BAIR), University of California, Berkeley}}

\twocolumn[{%
\renewcommand\twocolumn[1][]{#1}%
\maketitle
\begin{center}
    \centering
	\includegraphics[width=0.99\textwidth]{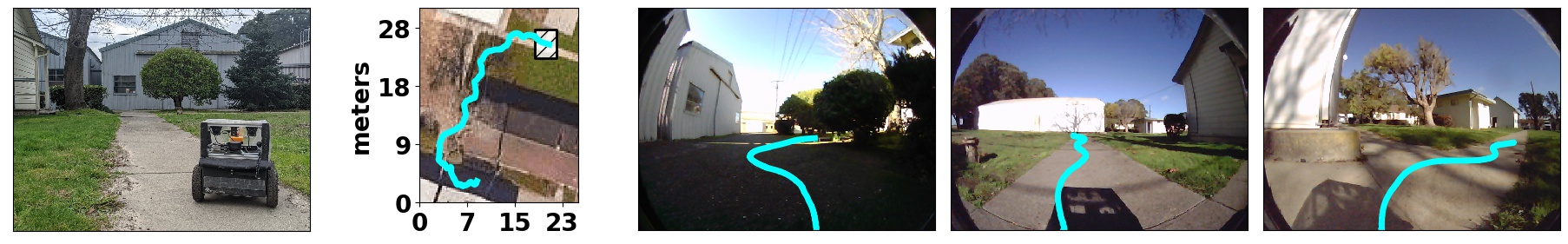}
	\includegraphics[width=0.99\textwidth]{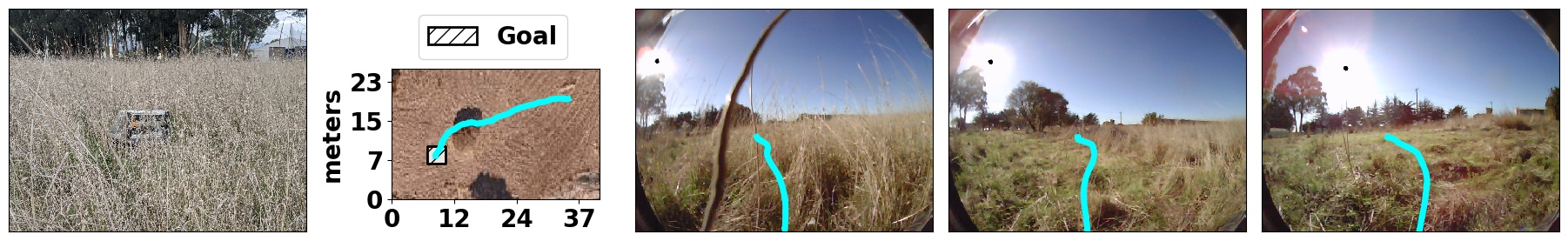}
	\captionof{figure}{\badgr is an end-to-end learning-based mobile robot navigation system that can be trained with self-supervised off-policy data gathered in real-world environments, without any simulation or human supervision. Using only RGB images and GPS, \badgr can follow sparse GPS waypoints without colliding while (top row) preferring smooth concrete paths and (bottom row) ignoring geometrically distracting obstacles such as tall grass.}
    \label{fig:teaser}
\end{center}%
}]

\begin{abstract}
Mobile robot navigation is typically regarded as a geometric problem, in which the robot's objective is to perceive the geometry of the environment in order to plan collision-free paths towards a desired goal. However, a purely geometric view of the world can can be insufficient for many navigation problems. For example, a robot navigating based on geometry may avoid a field of tall grass because it believes it is untraversable, and will therefore fail to reach its desired goal. In this work, we investigate how to move beyond these purely geometric-based approaches using a method that learns about physical navigational affordances from experience. Our approach, which we call \badgr, is an end-to-end learning-based mobile robot navigation system that can be trained with self-supervised off-policy data gathered in real-world environments, without any simulation or human supervision. \badgr can navigate in real-world urban and off-road environments with geometrically distracting obstacles. It can also incorporate terrain preferences, generalize to novel environments, and continue to improve autonomously by gathering more data. Videos, code, and other supplemental material are available on our website https://sites.google.com/view/badgr
\end{abstract}

\IEEEpeerreviewmaketitle


\section{Introduction}
\label{sec:intro}

Navigation for mobile robots is often regarded as primarily a geometric problem, where the aim is to construct either a local or global map of the environment, and then plan a path through this map~\citep{Thrun2008_book}. While this approach has produced excellent results in a range of settings, from indoor navigation~\citep{Rosen1968_SRI} to autonomous driving~\citep{Thorpe1988_TPAMI}, open-world mobile robot navigation often present challenges that are difficult to address with a purely geometric view of the world. Some geometric obstacles, such as the tall grass in Figure~\ref{fig:teaser}, may in fact be traversable. Different surfaces, though geometrically similar, may be preferred to differing degrees---for example, the vehicle in Fig.~\ref{fig:teaser} might prefer the paved surface over the bumpy field. Conventionally, these aspects of the navigation problem have been approached from the standpoint of \emph{semantic} understanding, using computer vision methods trained on human-provided traversability or road surface labels~\cite{Michels2005_ICML,Hadsell2009_JFR,Valada2017_ICRA,Hirose2018_IROS}. However, traversability, bumpiness, and other mobility-relevant attributes are \emph{physical} attributes of the environment, and in principle can be learned by the robot through experience, without human supervision. In this paper, we study how autonomous, self-supervised learning from experience can enable a robot to learn about the \emph{affordances} in its environment using raw visual perception and without human-provided labels or geometric maps.

Instead of using human supervision to teach a robot how to navigate, we investigate how the robot's own past experience can provide \emph{retrospective} self-supervision: for many physically salient navigational objectives, such as avoiding collisions or preferring smooth over bumpy terrains, the robot can autonomously measure how well it has fulfilled its objective, and then retrospectively label the preceding experience so as to learn a \emph{predictive} model for these objectives. For example, by experiencing collisions and bumpy terrain, the robot can learn, given an observation and a candidate plan of future actions, which actions might lead to bumpy or smooth to terrain, and which actions may lead to collision. This in effect constitutes a self-supervised multi-task reinforcement learning problem.

Based on this idea, we present a fully autonomous, self-improving, end-to-end learning-based system for mobile robot navigation, which we call \badgr---Berkeley Autonomous Driving Ground Robot. \badgr works by gathering off-policy data---such as from a random control policy---in real-world environments, and uses this data to train a model that predicts future relevant events---such as collision, position, or terrain properties---given the current sensor readings and the recorded executed future actions. Using this model, \badgr can then plan into the future and execute actions that avoid certain events, such as collision, and actively seek out other events, such as smooth terrain. \badgr constitutes a fully autonomous self-improving system because it gathers data, labels the data in a self-supervised fashion, trains the predictive model in order to plan and act, and can autonomously gather additional data to further improve itself.

While the particular components that we use to build \badgr draw on prior work~\citep{Kahn2018_ICRA,Kahn2018_CORL}, we demonstrate for the first time that a complete system based on these components can reason about both geometric and non-geometric navigational objectives, learn from off-policy data, does not require any human labelling, does not require expensive sensors or simulated data, can improve with autonomously collected data, and works in real-world environments.

The primary contribution of this work is an end-to-end learning-based mobile robot navigation system that can be trained entirely with self-supervised off-policy data gathered in real-world environments, without any simulation or human supervision. Our results demonstrate that our \badgr system can learn to navigate in real-world environments with geometrically distracting obstacles, such as tall grass, and can readily incorporate terrain preferences, such as avoiding bumpy terrain, using only 42 hours of autonomously collected data. Our experiments show that our method can outperform a LIDAR policy in complex real-world settings, generalize to novel environments, and can improve as it gathers more data.


\section{Related Work}
\label{sec:related}

Autonomous mobile robot navigation has been extensively studied in many real-world scenarios, ranging from indoor navigation~\citep{Rosen1968_SRI,How2008_control,Shen2011_ICRA,Paull2017_duckietown} to outdoor driving~\citep{Thorpe1988_TPAMI,Thrun2006_JFR,Scherer2008_IJRR,Urmson2008_NIPS,Furgale2010_JFR}. The predominant approach for autonomous navigation is to have the robot build a map, localize itself in the map, and use the map in order to plan and execute actions that enable the robot to achieve its goal. This simultaneous localization and mapping (SLAM) and planning approach~\citep{Thrun2008_book} has achieved impressive results, and underlies current state-of-the-art autonomous navigation technologies~\citep{Waymo,Skydio}. However, these approaches still have limitations, such as performance degradation in textureless scenes, requiring expensive sensors, and---most importantly---do not get better as the robot acts in the world~\citep{Fuentes2015_AIR}.

Learning-based methods have shown promise in addressing these limitations by learning from data. One approach to improve upon SLAM methods is to directly estimate the geometry of the scene~\citep{Fu2018_CVPR,Chang2018_CVPR,Wang2019_CVPR}. However, these methods are limited in that the geometry is only a partial description of the environment. Only learning about geometry can lead to unintended consequences, such as believing that a field of tall grass is untraversable. Semantic-based learning approaches attempt to address the limitations of purely geometric methods by associating the input sensory data with semantically meaningful labels, such as which pixels in an image correspond to traversable or bumpy terrain. However, these methods typically depend on existing SLAM approaches~\citep{Michels2005_ICML,Hadsell2009_JFR,Richter2017_RSS,Kahn2018_CORL,Wellhausen2019_RAL} or humans~\citep{Valada2017_ICRA,Hirose2018_IROS} in order to provide the semantic labels, which consequently means these approaches either inherit the limitations of geometric approaches or are not autonomously self-improving. Methods based on imitation learning have been demonstrated on real-world robots~\citep{Ross2013_ICRA,Bojarski2016_arxiv,Codevilla2018_ICRA}, but again depend on humans for expert demonstrations, which does not constitute a continuously self-improving system. End-to-end reinforcement learning approaches have shown promise in automating the entire navigation pipeline. However, these methods have typically focused on pure geometric reasoning, require on-policy data, and often utilize simulation due to constraints such as sample efficiency~\citep{Sadeghi2017_RSS,Kahn2018_ICRA,Savinov2018_ICLR,Bruce2018_CORL,Meng2019_ICRA,Chiang2019_RAL}. Prior works have investigated learning navigation policies directly from real-world experience, but typically require a person~\citep{Bruce2018_CORL,Loquercio2018_RAL,Hirose2019_RAL} or SLAM algorithm~\citep{Gandhi2017_IROS} to gather the data, assume access to the ground-truth robot state~\citep{Riedmiller2007_FBIT}, learn using low-bandwidth sensors~\citep{Mahmood2018_CoRL}, or only perform collision avoidance~\citep{Kahn2018_ICRA,Kendall2019_ICRA}. Our approach overcomes the limitations of these prior works by designing an end-to-end reinforcement learning approach that directly learns to predict relevant navigation cues with a sample-efficient, off-policy algorithm, and can continue to improve with additional experience via a self-supervised data labelling mechanism that does not depend on humans or SLAM algorithms.

The most similar works to our \badgr system are GCG~\citep{Kahn2018_ICRA} and CAPs~\citep{Kahn2018_CORL}. However, GCG only learned to avoid collisions and CAPs required human supervision in order to learn non-collision avoidance behaviors, while \badgr is able to reason about both geometric and non-geometric navigational without any human supervision in complex, real-world environments.


\section{Berkeley Autonomous Driving Ground Robot}
\label{sec:method}

Our goal is to enable a mobile robot to navigate in real-world environments. We therefore developed \badgr, an end-to-end learning-based mobile robot navigation system that can be trained entirely with self-supervised, off-policy data gathered in real-world environments, without any simulation or human supervision, and can improve as it gathers more data.

\badgr operates by autonomously gathering large amounts of off-policy data in real-world environments. Using this data, \badgr labels relevant events---such as collisions or bumpy terrain---in a self-supervised manner, and adds these labelled events back into the dataset. \badgr then trains a predictive model that takes as input the current observation, such as camera images, and a future sequence of actions, such as linear and angular velocity commands, and predicts the relevant future events. When deploying the trained \badgr system, the user designs a reward function that encodes the task they want the robot to accomplish in terms of these relevant events---such as to reach a goal while avoiding collisions and bumpy terrain---and the robot autonomously plans and executes actions that maximize this reward.

In order to build a self-supervised learning-based navigational system, we must assume that \emph{retrospective} supervision is available to the robot. This means that the robot must be able to experience events, such as collisions, and then learn to avoid (or seek out) such events in the future. We therefore make the following assumptions: (1) the robot can only learn about events it has experienced and that can be measured using the onboard sensors and (2) experiencing these events, even undesirable events such as colliding, is acceptable. We believe these assumptions are realistic for many real-world autonomous mobile robot applications.

In the following sections, we will describe the robot, data collection and labelling, model training, and planning components, followed by a summarizing overview of the entire system.


\subsection{Mobile Robot Platform}
\label{sec:method-robot}

The specific design considerations for the robotic platform focus on enabling long-term autonomy with minimal human intervention.

\begin{wrapfigure}{r}{0.5\columnwidth}
    \vspace*{-10pt}
    \centering
    \includegraphics[width=0.5\columnwidth,trim={0 0 0 0}, clip]{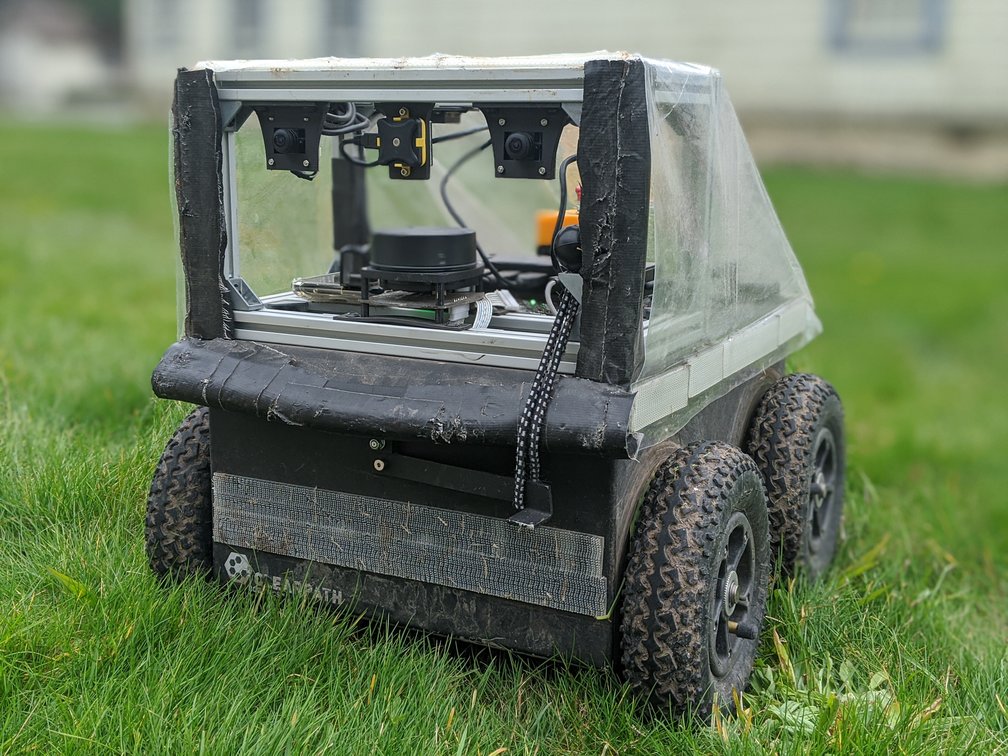}
    \caption{The mobile robot platform.}
    \label{fig:method-jackal}
    \vspace*{-10pt}
\end{wrapfigure}%
The robot we use is the Clearpath Jackal, shown in Fig.~\ref{fig:method-jackal}. The Jackal measures $508\text{mm} \times 430\text{mm} \times 250\text{mm}$ and weighs 17kg, making it ideal for navigating in both urban and off-road environments. The Jackal is controlled by specifying the desired linear and angular velocity, which are used as setpoints for the low-level differential drive controllers. The default sensor suite consists of a 6-DOF IMU, which measures linear acceleration and angular velocity, a GPS unit for approximate global position estimates, and encoders to measure the wheel velocity. In addition, we added the following sensors on top of the Jackal: two forward-facing $170\degree$ field-of-view $640 \times 480$ cameras, a 2D LIDAR, and a compass.

Inside the Jackal is an NVIDIA Jetson TX2 computer, which is ideal for running deep learning applications, in addition to interfacing with the sensors and low-level microcontrollers. Data is saved to an external SSD, which must be large and fast enough to store 1.3GB per minute streaming in from the sensors. Experiments were monitored remotely via a 4G smartphone mounted on top of the Jackal, which allowed for video streaming and, if needed, teleoperation.

The robot was designed primarily for robustness, with a relatively minimal and robust sensor suite, in order to enable long-term autonomous operation for large-scale data collection.


\subsection{Data Collection}
\label{sec:method-data}

We design the data collection methodology to enable gathering large amounts of diverse data for training with minimal human intervention. 

The first consideration when designing the data collection policy is whether the learning algorithm requires on-policy data. On-policy data collection entails alternating between gathering data using the current policy, and retraining the policy using the most recently gathered data. On-policy data collection is highly undesirable because only the most recently gathered data can be used for training; all previously gathered data must be thrown out. In contrast, off-policy learning algorithms can train policies using data gathered by any control policy. Due to the high cost of gathering data with real-world robotic systems, we choose to use an off-policy learning algorithm in order to be able to gather data using any control policy and train on all of the gathered data.

\begin{figure}[H]
    \centering
    \includegraphics[width=0.8\columnwidth]{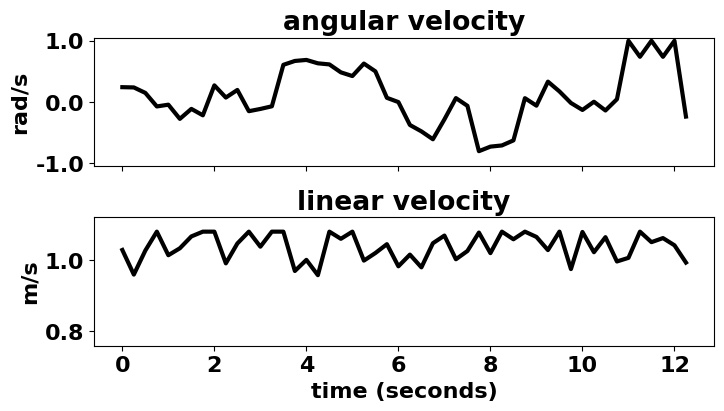}
    \caption{An example plot of the commanded angular and linear velocities from our time-correlated random control policy that is used to gather data.}
    \label{fig:method-noise}
\end{figure}

The second consideration when designing the data collection policy is to ensure the environment is sufficiently explored, while also ensuring that the robot execute action sequences it will realistically wish to execute at test time. A na\"ive uniform random control policy is inadequate because the robot will primarily drive straight due to the linear and angular velocity action interface of the robot, which will result in both insufficient exploration and unrealistic test time action sequences. We therefore use a time-correlated random walk control policy to gather data, which is visualized in Fig.~\ref{fig:method-noise}.

\begin{wrapfigure}{r}{0.4\columnwidth}
    \vspace*{-15pt}
    \centering
    \includegraphics[width=0.4\columnwidth,trim={0 0 0 4cm}, clip]{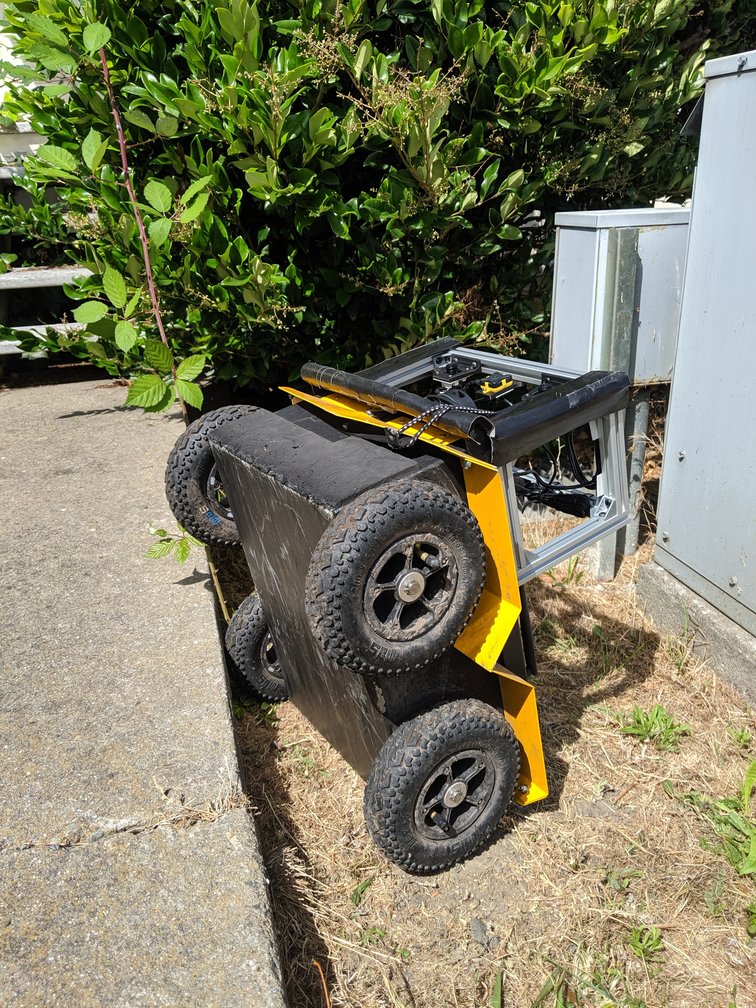}
    \caption{\footnotesize{While collecting data, the robot will periodically require a manual intervention to reset from catastrophic failures, though recovery is usually automatic.}}
    \label{fig:method-jackal-flipped}
    \vspace*{-10pt}
\end{wrapfigure}%
As the robot is gathering data using the random control policy, it will require a mechanism to detect if it is in collision or stuck, and an automated controller to reset itself in order to continue gathering data. We detect collisions in one of two ways, either using the LIDAR to detect when an obstacle is near or the IMU to detect when the robot is stuck due to an obstacle. We used the LIDAR collision detector in urban environments in order to avoid damaging property, and the IMU collision detector in off-road environments because the LIDAR detector was overly pessimistic, such as detecting grass as an obstacle. Once a collision is detected, a simple reset policy commands the robot to back up and rotate. However, sometimes the reset policy is insufficient, for example if the robot flips over (Fig.~\ref{fig:method-jackal-flipped}), and a person must manually reset the robot.

As the robot collects data, all the raw sensory data is saved onboard. After data collection for the day is complete, the data is copied to a desktop machine and subsampled down to 4Hz.


\subsection{Self-Supervised Data Labelling}
\label{sec:method-data-labelling}

\badgr then goes through the raw, subsampled data and calculates labels for specific navigational events. These events consist of anything pertinent to navigation that can be extracted from the data in a self-supervised fashion.

In our experiments, we consider three different events: collision, bumpiness, and position. A collision event is calculated as occurring when, in urban environments, the LIDAR measures an obstacle to be close or, in off-road environments, when the IMU detects a sudden drop in linear acceleration and angular velocity magnitudes. A bumpiness event is calculated as occurring when the angular velocity magnitudes measured by the IMU are above a certain threshold. The position is determined by an onboard state estimator that fuses wheel odometry and the IMU to form a local position estimate.

After \badgr has iterated through the data, calculated the event labels at each time step, and added these event labels back into the dataset, \badgr can then train a model to predict which actions lead to which navigational events.


\subsection{Predictive Model}
\label{sec:method-model}

The learned predictive model takes as input the current sensor observations and a sequence of future intended actions, and predicts the future navigational events. We denote this model as $f_\theta(\bo_t, \ba_{t:t+H}) \rightarrow \hat{\be}^{0:K}_{t:t+H}$, which defines a function $f$ parameterized by vector $\theta$ that takes as input the current observation $\bo_t$ and a sequence of $H$ future actions ${\ba_{t:t+H} = (\ba_t, \ba_{t+1}, ..., \ba_{t+H-1})}$, and predicts $K$ different future events ${\hat{\be}^k_{t:t+H} = (\hat{\be}^k_t, \hat{\be}^k_{t+1}, ..., \hat{\be}^k_{t+H-1}) \;\forall k \in \{0, ..., K-1\}}$.

\begin{figure*}[t]
	\centering
	\includegraphics[width=\textwidth]{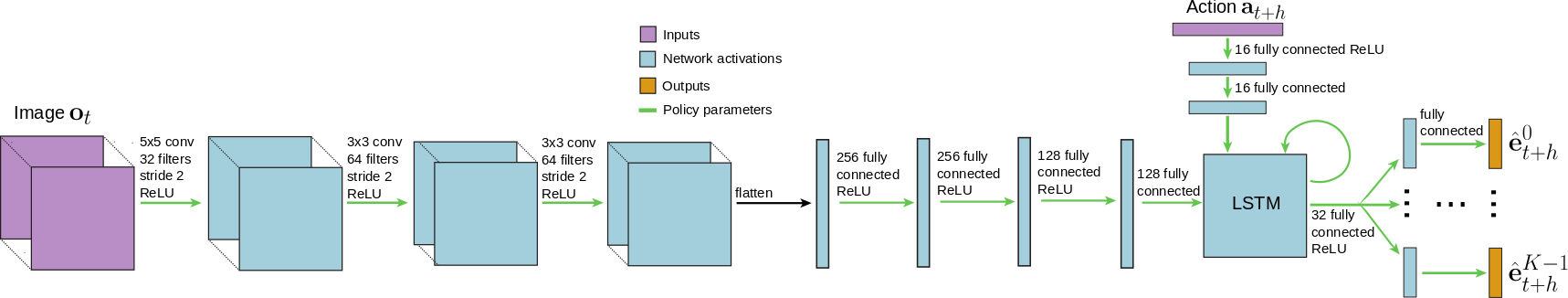}
	\caption{Illustration of the deep neural network predictive model at the core of our learning-based navigation policy. The neural network takes as input the current RGB image and processes it with convolutional and fully connected layers to form the initial hidden state of a recurrent LSTM unit~\citep{Hochreiter1997_neural}. This recurrent unit takes as input $H$ actions in a sequential fashion, and produces $H$ outputs. These outputs of the recurrent unit are then passed through additional fully connected layers to predict all $K$ events for all $H$ future time steps. These predicted future events, such as position, if the robot collided, and if the robot drove over bumpy terrain, enable a planner to select actions that achieve desirable events, such reaching a goal, and avoid undesirable events, such as collisions and bumpy terrain.}
	\label{fig:method-nn}
\end{figure*}%

The model we learn is an image-based, action-conditioned predictive deep neural network, shown in Fig.~\ref{fig:method-nn}. The network first processes the input image observations using convolutional and fully connected layers. The final output of the these layers serves as the initialization for a recurrent neural network, which sequentially processes each of the $H$ future actions $\ba_{t+h}$ and outputs the corresponding predicted future events $\hat{\be}^{0:K}_{t+h}$.

The model is trained---using the observations, actions, and event labels from the collected dataset---to minimize a loss function that penalizes the distance between the predicted and ground truth events
\begin{align}
    \loss(\theta, \dataset) = \sum_{(\bo_t, \ba_{t:t+H}) \in \dataset} \sum_{k=0}^{K-1} \loss^k(\hat{\be}^k_{t:t+H}, \be^k_{t:t+H}) : \label{eqn:method-training-loss} \\
    \hat{\be}^{0:K}_{t:t+H} = f_\theta(\bo_t, \ba_{t:t+H}). \nonumber
\end{align}
The individual losses $\loss^k$ for each event are either cross entropy if the event is discrete, or mean squared error if the event is continuous. The neural network parameter vector $\theta$ is trained by performing minibatch gradient descent on the loss in Eqn.~\ref{eqn:method-training-loss}.


\subsection{Planning}
\label{sec:method-planning}

Given the trained neural network predictive model, this model can then be used at test time to plan and execute desirable actions.

We first define a reward function $R(\hat{\be}^{0:K}_{t:t+H})$ that encodes what we want the robot to do in terms of the model's predicted future events. For example, the reward function could encourage driving towards a goal while discouraging collisions or driving over bumpy terrain. The specific reward function we use is specified in the experiments section.

Using this reward function and the learned predictive model, we solve the following planning problem at each time step
\begin{align}
\ba^*_{t:t+H} = \arg\max_{\ba_{t:t+H}} R(f_\theta(\bo_t, \ba_{t:t+H})), \label{eqn:method-planning-opt}
\end{align}
execute the first action, and continue to plan and execute following the framework of model predictive control.

We solve Eqn.~\ref{eqn:method-planning-opt} using the zeroth order stochastic optimizer from~\citet{Nagabandi2019_CoRL}. This optimizer works by maintaining a running estimate $\hat{\ba}_{0:H}$ of the optimal action sequence. Each time the planner is called, $N$ action sequences $\tilde{\ba}^{0:N}_{0:H}$ are sampled that are time-correlated and centered around this running action sequence estimate
\begin{align}
\epsilon^n_h &\sim \mathcal{N}(0, \sigma \cdot \mathbf{I}) \; \forall n \in \{0...N-1\}, h \in \{0...H-1\} \label{eqn:method-planning-sampling} \\
\tilde{\ba}^n_h &= \beta \cdot (\hat{\ba}_{h+1} + \epsilon^n_h) + (1 - \beta) \cdot \tilde{\ba}^n_{h-1} \text{ where } \tilde{\ba}_{h<0} = 0, \nonumber
\end{align}
in which the parameter $\sigma$ determines how close the sampled action sequences should be to the running action sequence estimate, and the parameter $\beta \in [0, 1]$ determines the degree to which the sampled action sequences are correlated in time.

Each action sequence is then propagated through the predictive model in order to calculate the reward ${\tilde{R}^n = R(f_\theta(\bo_t, \tilde{\ba}^n_{0:H}))}$. Given each action sequence and its corresponding reward, we update the running estimate of the optimal action sequence via a reward-weighted average
\begin{align}
\hat{\ba}_{0:H} &= \frac{\sum_{n=0}^N \exp(\gamma \cdot R^n) \cdot \tilde{\ba}^n_{0:H}}{\sum_{n'=0}^N \exp(\gamma \cdot R^{n'})}, \label{eqn:method-planning-update}
\end{align}
in which $\gamma \in \mathrm{R}^+$ is a parameter that determines how much weight should be given to high-reward action sequences.

Each time the planner is called, new action sequences are sampled according to Eqn.~\ref{eqn:method-planning-sampling}, these action sequences are propagated through the learned predictive model in order to calculate the reward of each sequence, the running estimate of the optimal action sequence is updated using Eqn.~\ref{eqn:method-planning-update}, and the robot executes the first action $\hat{\ba}_0$.

This optimizer is more powerful than other zeroth order stochastic optimizers, such as random shooting or the cross-entropy method~\citep{Rubinstein2013_springer}, because it warm-starts the optimization using the solution from the previous time step, uses a soft update rule for the new sampling distribution in order to leverage all of the sampled action sequences, and considers the correlations between time steps. In our experiments, we found this more powerful optimizer was necessary to achieve good planning.


\subsection{Algorithm Summary}
\label{sec:method-summary}

We now provide a brief summary of how our \badgr system operates during training (Alg.~\ref{alg:train}) and deployment (Alg.~\ref{alg:test}).

During training, \badgr gathers data by executing actions according to the data collection policy and records the onboard sensory observations and executed actions. Next, \badgr uses the gathered dataset to self-supervise the event labels, which are added back into the dataset. This dataset is then used to train the learned predictive model.

When deploying \badgr, the user first defines a reward function that encodes the specific task they want the robot to accomplish. \badgr then uses the trained predictive model, current observation, and reward function to plan a sequence of actions that maximize the reward function. The robot executes the first action in this plan, and \badgr continues to alternate between planning and executing until the task is complete.

\begin{algorithm}[!h]
\caption{Training \badgr}
\label{alg:train}
\begin{algorithmic}[1]
\STATE initialize dataset $\mathcal{D} \leftarrow \emptyset$
\WHILE{not done collecting data}
	\STATE get current observation $\bo_t$ from sensors
	\STATE get action $\ba_t$ from data collection policy
	\STATE add $(\bo_t, \ba_t)$ to $\mathcal{D}$
	\STATE execute $\ba_t$
	\IF{in collision}
		\STATE execute reset maneuver
	\ENDIF
\ENDWHILE
\FOR{\textbf{each} $(\bo_t, \ba_t) \in \mathcal{D}$}
	\STATE calculate event labels $\be_t^{0:K}$ using self-supervision
	\STATE add $\be_t^{0:K}$ to $\mathcal{D}$
\ENDFOR
\STATE use $\mathcal{D}$ to train predictive model $f_\theta$ by minimizing Eqn.~\ref{eqn:method-training-loss}
\end{algorithmic}
\end{algorithm}

\begin{algorithm}[!h]
\caption{Deploying \badgr}
\label{alg:test}
\begin{algorithmic}[1]
\STATE \textbf{input}: trained predictive model $f_\theta$, reward function $R$
\WHILE{task is not complete}
	\STATE get current observation $\bo_t$ from sensors
	\STATE solve Eqn.~\ref{eqn:method-planning-opt} using $f_\theta, \bo_t,$ and $R$ \\to get the planned action sequence $\ba^*_{t:t+H}$
	\STATE execute the first action $\ba^*_t$
\ENDWHILE
\end{algorithmic}
\end{algorithm}


\section{Experiments}
\label{sec:exp}

In our experimental evaluation, we study how \badgr can autonomously learn a successful navigation policy in real-world environments, improve as it gathers more data, generalize to unseen environments, and compare it to purely geometric approaches. Videos, code, and other supplemental material are available on our website\footnote{\url{https://sites.google.com/view/badgr}}.

We performed our evaluation in a real-world outdoor environment consisting of both urban and off-road terrain. \badgr autonomously gathered 34 hours of data in the urban terrain and 8 hours in the off-road terrain. Although the amount of data gathered may seem significant, the total dataset consisted of 720,000 off-policy datapoints, which is smaller than currently used datasets in computer vision~\cite{Deng2009_CVPR} and significantly smaller than the number of samples often used by deep reinforcement learning algorithms~\cite{Hessel2018_AAAI}.

Our evaluations consist of tasks that involve reaching a goal GPS location, avoiding collisions, and preferring smooth over bumpy terrain. In order for \badgr to accomplish these tasks, we design the reward function that \badgr uses for planning as such
\begin{align}
&R(\hat{\be}^{0:K}_{t:t+H}) = -\sum_{t'=t}^{t+H-1} R^{\textsc{coll}}(\hat{\be}_{t'}^{0:K})  + \label{eqn:method-planning-reward} \\
&\hspace{94pt}\alpha^{\textsc{pos}} \cdot R^{\textsc{pos}}(\hat{\be}_{t'}^{0:K}) + \alpha^{\textsc{bum}} \cdot R^{\textsc{bum}}(\hat{\be}_{t'}^{0:K}) \nonumber \\
&R^{\textsc{coll}}(\hat{\be}_{t'}^{0:K}) = \;\hat{\be}^{\textsc{coll}}_{t'} \nonumber \\
&R^{\textsc{pos}}(\hat{\be}_{t'}^{0:K}) = \;(1 - \hat{\be}^{coll}_{t'}) \cdot \frac{1}{\pi} \angle(\hat{\be}^{\textsc{pos}}_{t'}, \mathbf{p}^{\textsc{goal}}) + \hat{\be}^{coll}_{t'} \nonumber \\
&R^{\textsc{bum}}(\hat{\be}_{t'}^{0:K}) = \;(1 - \hat{\be}^{coll}_{t'}) \cdot \hat{\be}^{\textsc{bum}}_{t'} + \hat{\be}^{coll}_{t'}, \nonumber
\end{align}
where $\alpha^{\textsc{pos}}$ and $\alpha^{\textsc{bum}}$ are user-defined scalars that weight how much the robot should care about reaching the goal and avoiding bumpy terrain. An important design consideration for the reward function was how to encode this multi-objective task. First, we ensured each of the individual rewards were in the range of $[0, 1]$, which made it easier to weight the individual rewards. Second, we ensured the collision reward always dominated the other rewards: if the robot predicted it was going to collide, all of the individual rewards were assigned to their maximum value of 1; conversely, if the robot predicted it was not going to collide, all of the individual rewards were assigned to their respective values.

\begin{figure}[t]
    \centering
    \captionsetup[subfigure]{aboveskip=1pt,belowskip=5pt}
    \begin{subfigure}[t]{\columnwidth}
	    \includegraphics[width=\columnwidth]{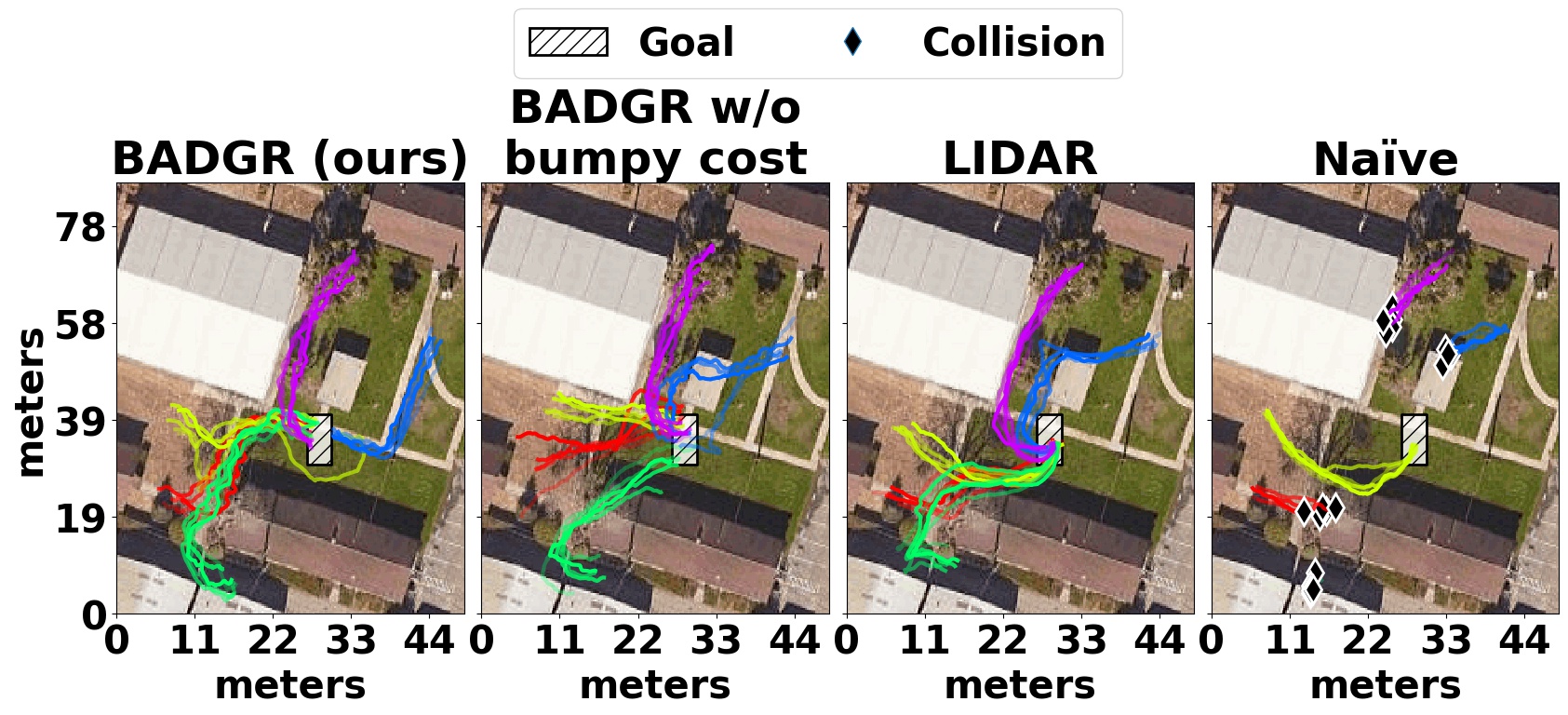}
	\end{subfigure}
	\resizebox{\columnwidth}{!}{%
	\begin{tabular}{|l|c|c|}
		\hline Method & Successfully Reached Goal & Avg. Bumpiness \\
		\hline \hline \textbf{\badgr (ours)} & \textbf{25/25 (100\%)} & $\mathbf{8.7 \pm 4.4}$  \\
		\hline \badgr w/o bumpy cost & 25/25 (100\%) & $15.0 \pm 3.4$ \\
		\hline LIDAR & 25/25 (100\%) & $13.3 \pm 2.9$ \\
		\hline Na\"ive & 5/25 (20\%) & N/A \\
		\hline
	\end{tabular}}
    \caption{Experimental evaluation in an urban environment for the task of reaching a specified goal position while avoiding collisions and bumpy terrain. Each approach was evaluated from 5 different start locations---each color corresponding to a different start location---with 5 runs per each start location. The figures show the paths of each run, and whether the run successfully reached the goal or ended in a collision. The table shows the success rate and average bumpiness for each method. Our \badgr approach is better able to reach the goal and avoid bumpy terrain compared to the other methods.}
    \label{fig:results-urban}
\end{figure}

\begin{figure}[t]
    \centering
    \begin{subfigure}[t]{\columnwidth}
    	\includegraphics[height=0.114\textheight,trim={27cm 2cm 45cm 2cm},clip]{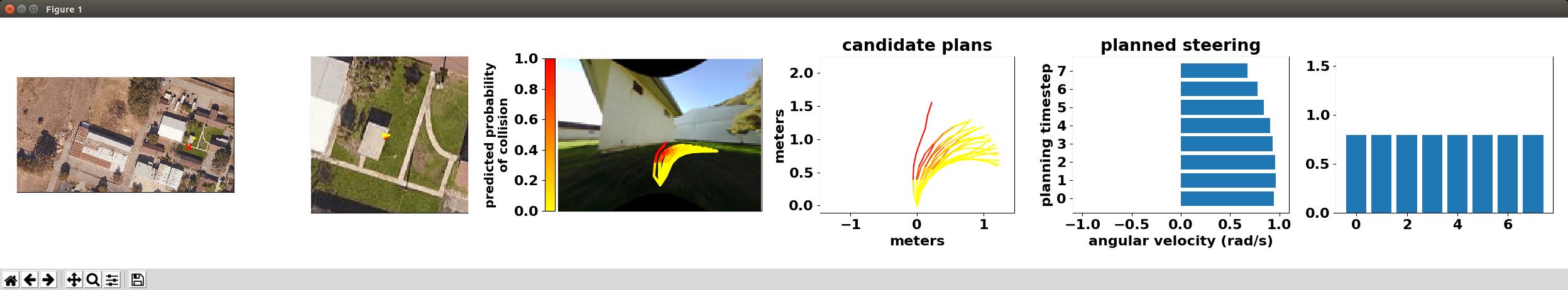}
    	\hfill
    	\includegraphics[height=0.114\textheight,trim={31.3cm 2cm 45cm 2cm},clip]{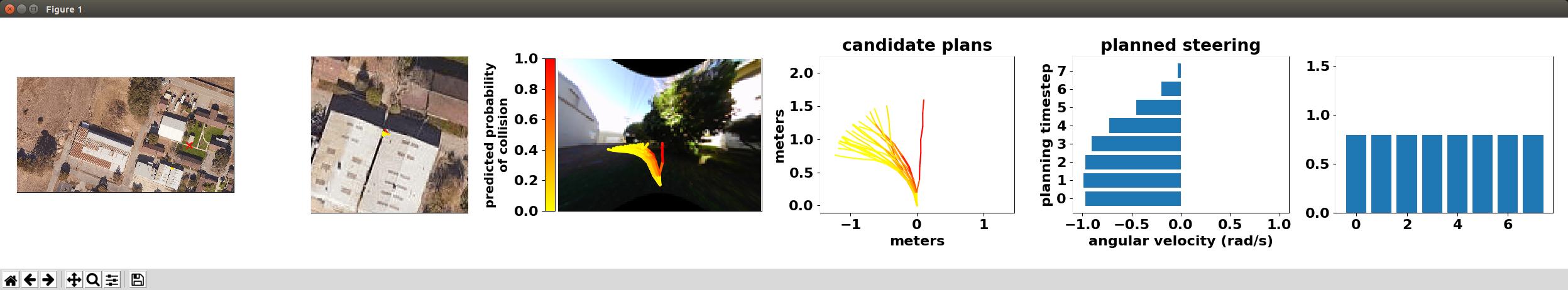}
    	\hfill
    	\includegraphics[height=0.114\textheight,trim={31.3cm 2cm 45cm 2cm},clip]{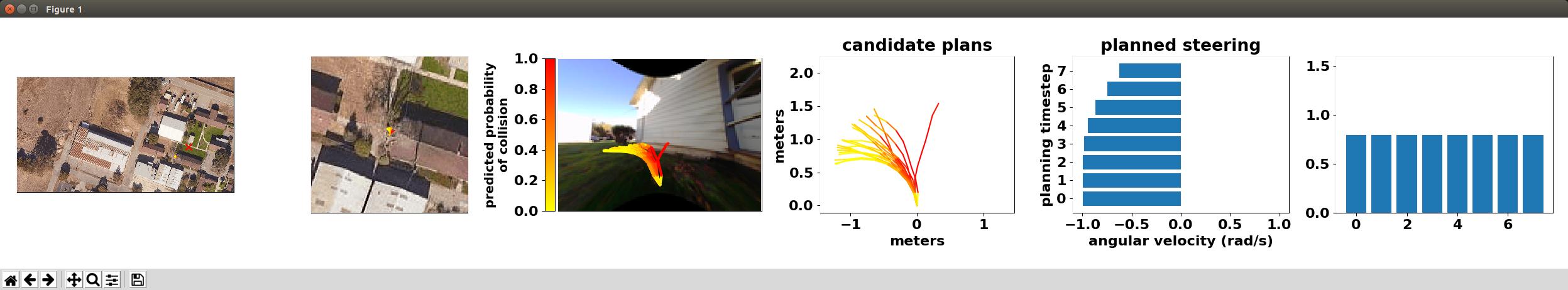}
    \end{subfigure}
    \par\vspace{-10pt}
    \begin{subfigure}[t]{\columnwidth}
    	\includegraphics[height=0.114\textheight,trim={27cm 2cm 45cm 2cm},clip]{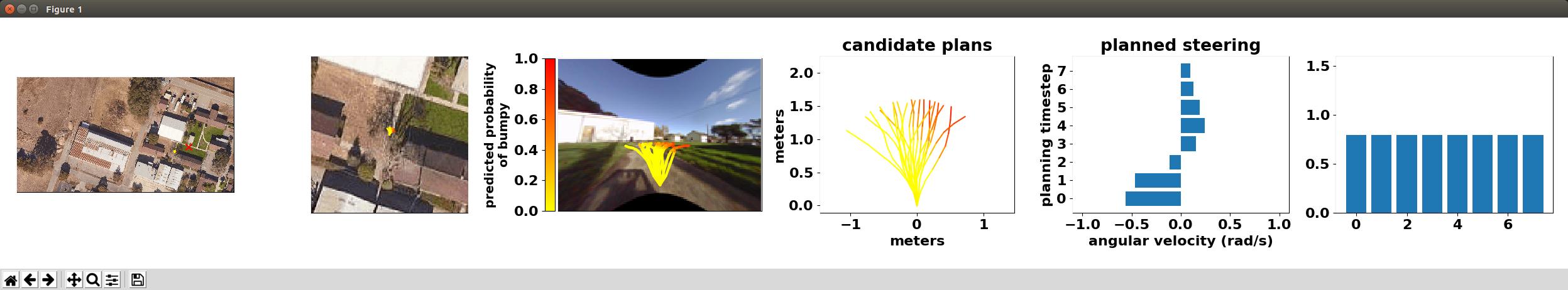}
	\hfill
	    \includegraphics[height=0.114\textheight,trim={31.3cm 2cm 45cm 2cm},clip]{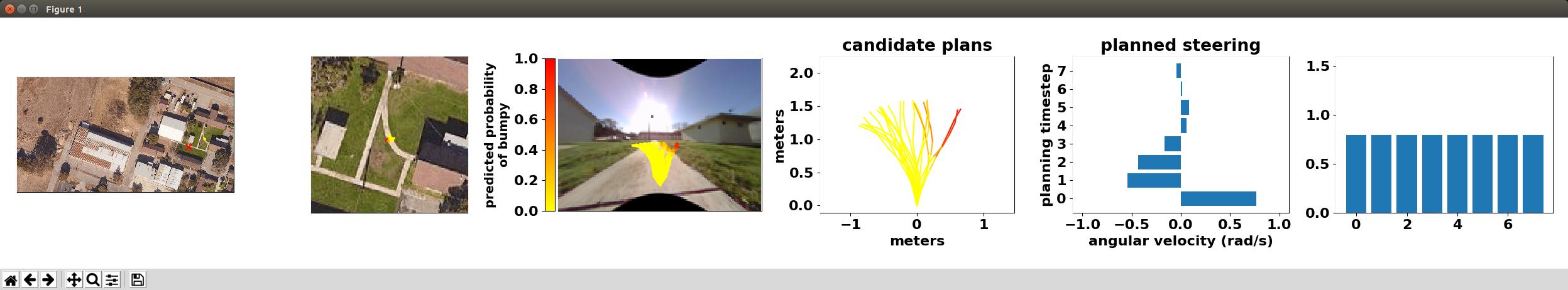}
	\hfill
	    \includegraphics[height=0.114\textheight,trim={31.3cm 2cm 45cm 2cm},clip]{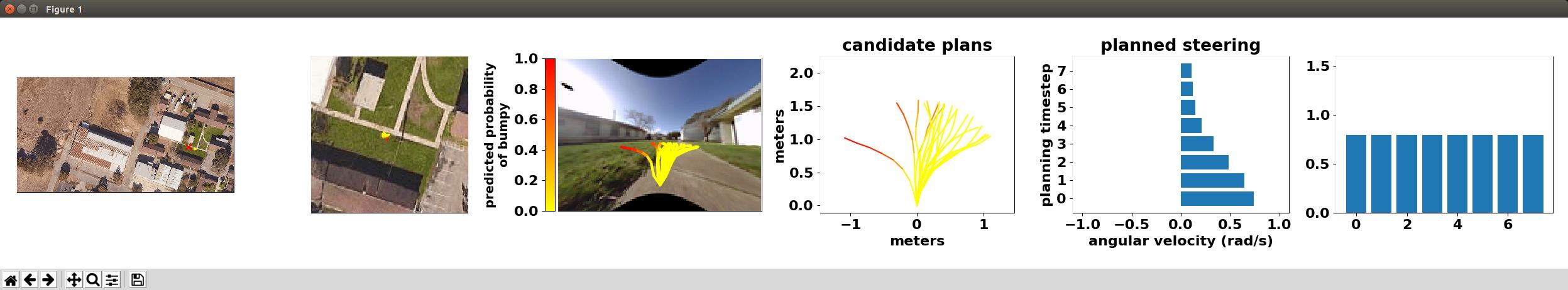}
    \end{subfigure}
	\caption{Visualization of \badgr 's  predictive model in the urban environment. Each image shows the candidate paths that \badgr considers during planning. These paths are color coded according to either their probability of collision (top row) or probability of experiencing bumpy terrain (bottom row) according to {\badgr}'s learned predictive model. These visualizations show the learned model can accurately predict that action sequences which would drive into buildings or bushes will result in a collision, and that action sequences which drive on concrete paths are smoother than driving on grass.}
    \label{fig:results-actionselection-urban}
\end{figure}

We evaluated \badgr against two other methods:
\begin{itemize}
	\item \textit{LIDAR}: a policy that drives towards the goal while avoiding collisions using the range measurements from an onboard 2D LIDAR. Note that our method only uses the camera images, while this approach uses LIDAR.
	\item \textit{Na\"ive}: a na\"ive policy that simply drives straight towards the specified goal.
\end{itemize}
We compare against LIDAR, which is a common geometric-based approach for designing navigation policies, in order to demonstrate the advantages of our learning-based approach, while the purpose of the na\"ive policy is to provide a lower bound baseline and calibrate the difficulty of the task.

Note that for all tasks, only a single GPS coordinate---the location of the goal---is given to the robot. This single GPS coordinate is insufficient for successful navigation, and therefore the robot must use other onboard sensors in order to accomplish the task.

\textbf{Urban environment.}
We first evaluated all the approaches for the task of reaching a goal GPS location while avoiding collisions and bumpy terrain in an urban environment.
Fig.~\ref{fig:results-urban} shows the resulting paths that \badgr, LIDAR, and the na\"ive policies followed. The na\"ive policy almost always crashed, which illustrates the urban environment contains many obstacles. The LIDAR policy always succeeded in reaching the goal, but failed to avoid the bumpy grass terrain. \badgr also always succeeded in reaching the goal, and---as also shown by Fig.~\ref{fig:results-urban}---succeeded in avoiding bumpy terrain by driving on the paved paths. Note that we never told the robot to drive on paths; \badgr automatically learned from the onboard camera images that driving on concrete paths is smoother than driving on the grass.

While a sufficiently high-resolution 3D LIDAR could in principle identify the bumpiness of the terrain and detect the paved paths automatically, 3D geometry is not a perfect indicator of the terrain properties. For example, let us compare tall grass versus gravel terrain. Geometrically, the tall grass is  bumpier than the gravel, but when actually driving over these terrains, the tall grass will result in a smoother ride. This example underscores the idea that there is not a clear mapping between geometry and physically salient properties such as whether terrain is smooth or bumpy.

\badgr overcomes this limitation by directly learning about physically salient properties of the environment using the raw onboard observations---in this case, the IMU readings---to determine if the terrain is bumpy. Our approach does not make assumptions about geometry, but rather lets the predictive model learn correlations from the onboard sensors; Fig.~\ref{fig:results-actionselection-urban} shows our predictive model successfully learns which image and action sequences lead to collisions and bumpy terrain and which do not.

\begin{figure}[t]
    \centering
    \includegraphics[width=\columnwidth]{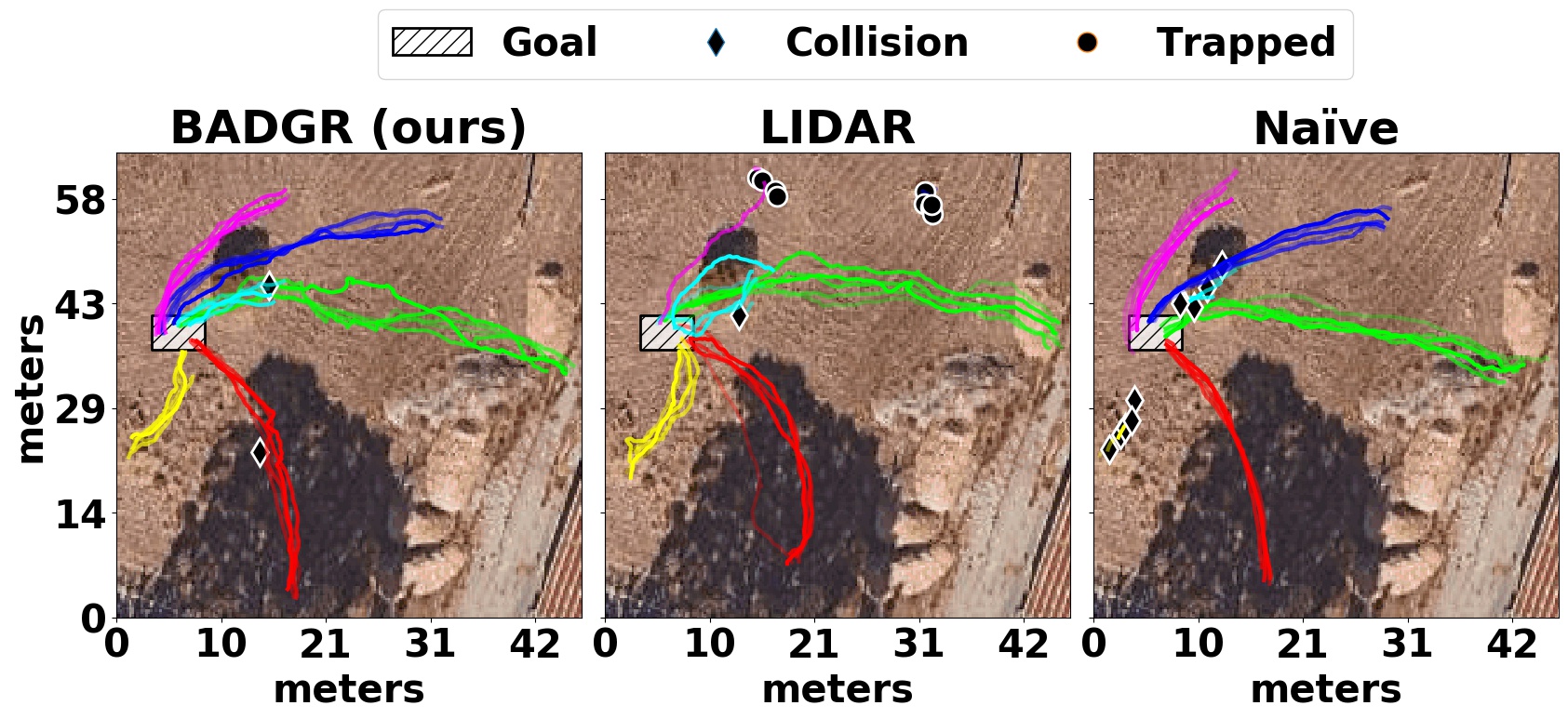}
	\caption{Experimental evaluation in an off-road environment for the task of reaching a specified goal location while avoiding collisions. Each approach was evaluated from 5 different start locations---each color corresponding to a different start location---with 5 runs per each start location. Each run terminated when the robot collided, failed to make progress and was trapped, or successfully reached the goal. Our \badgr policy is the only approach which can consistently reach the goal without colliding or getting trapped.}
    \label{fig:results-off-road-paths}
\end{figure}

\begin{figure}[t]
	\centering
	\begin{subfigure}[c]{0.07\columnwidth}
	\caption{}\label{fig:results-off-road-tpv-images-lidar}
	\end{subfigure}%
	\begin{minipage}[c]{0.93\columnwidth}
	\includegraphics[width=0.322\columnwidth]{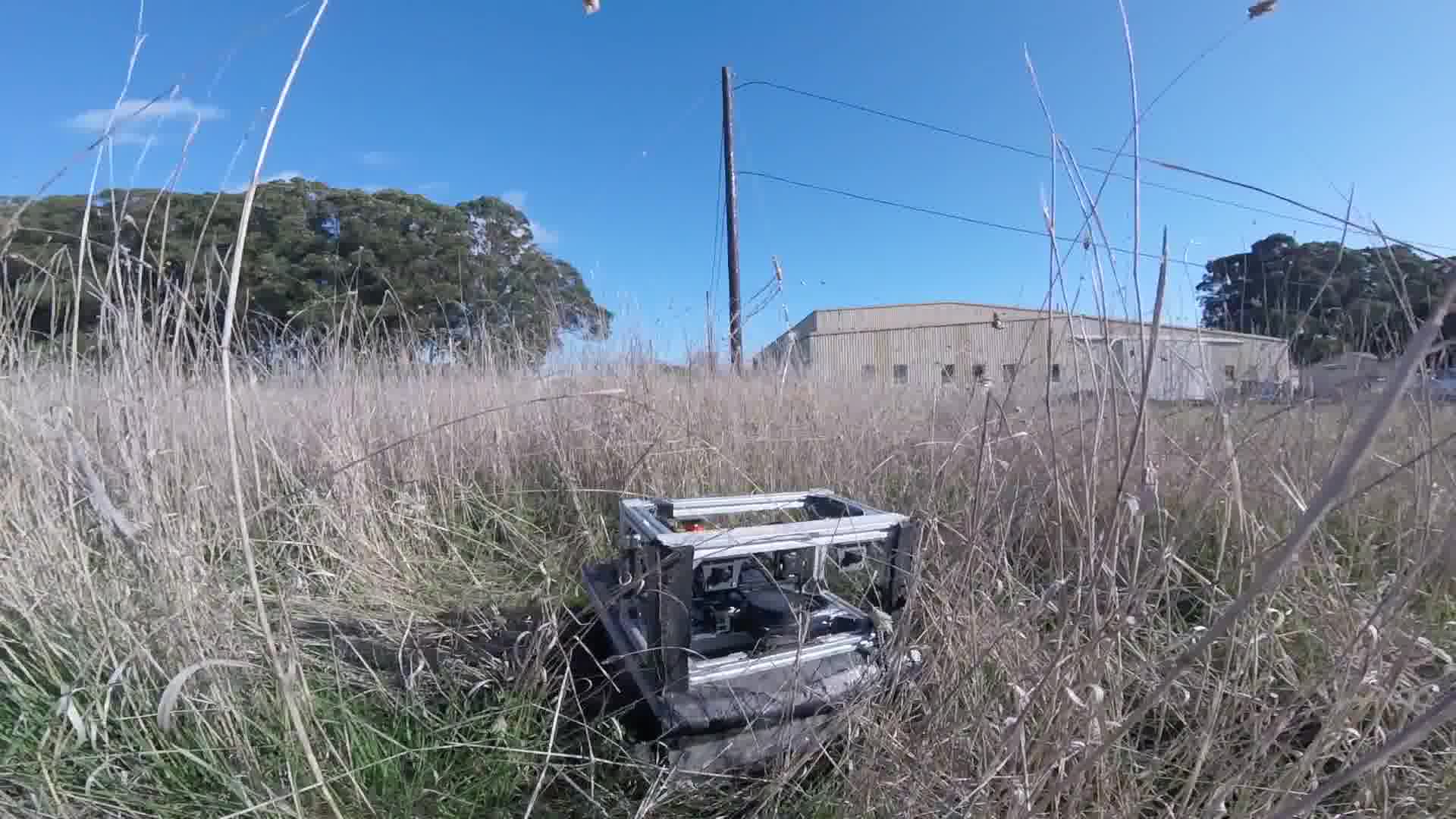}
	\hfill
	\includegraphics[width=0.322\columnwidth]{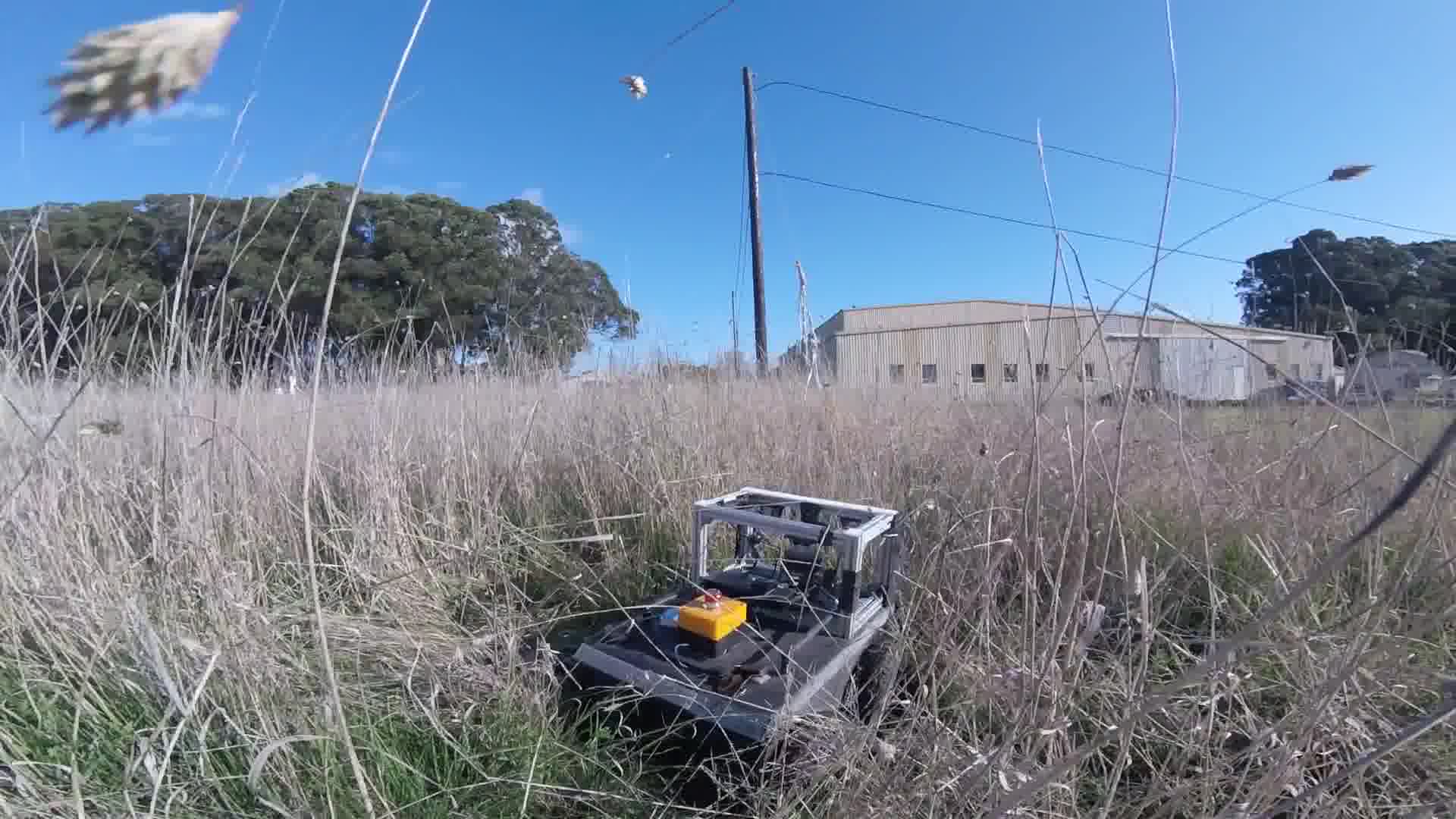}
	\hfill
	\includegraphics[width=0.322\columnwidth]{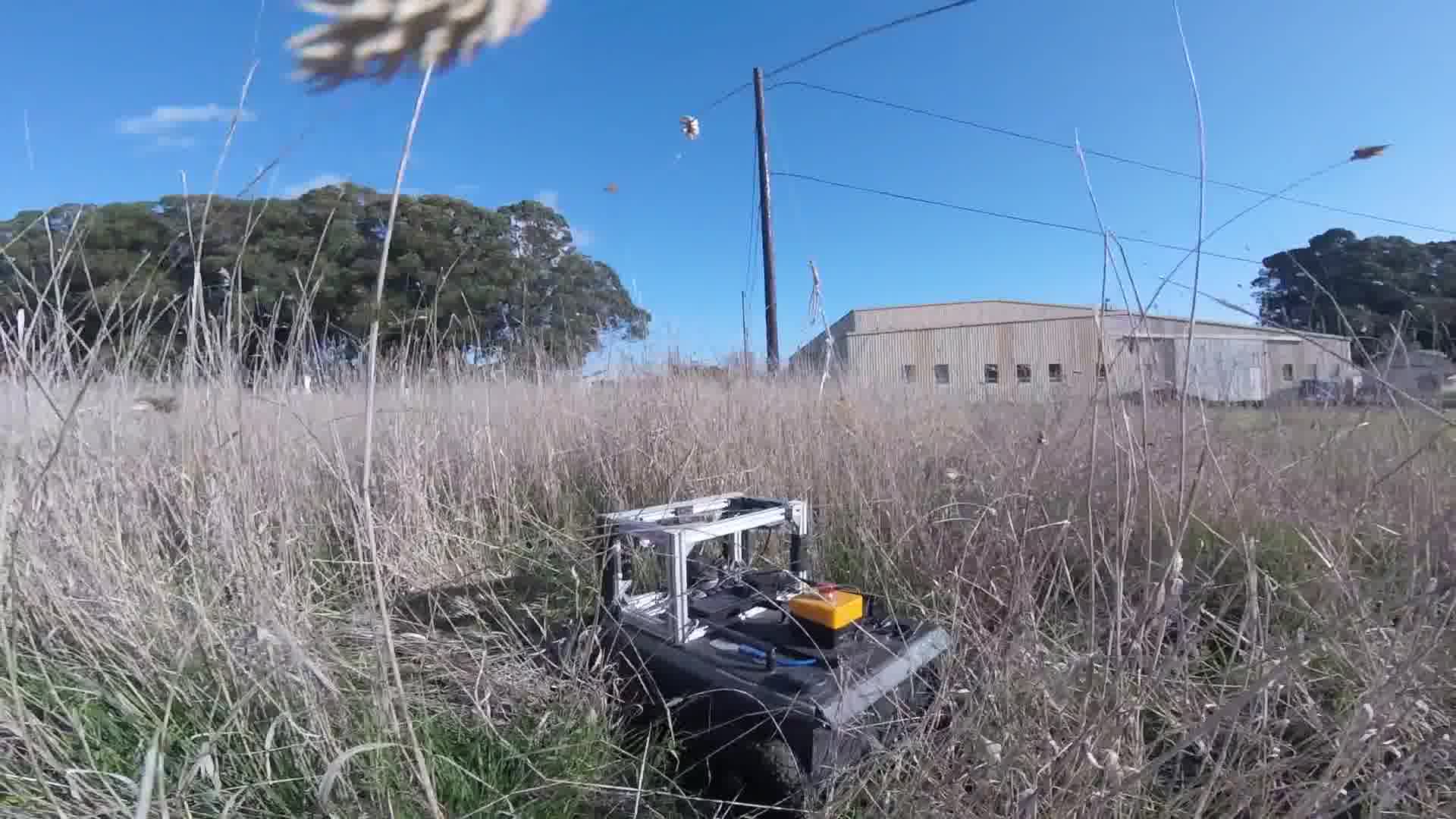}
	\end{minipage}
	\par\vspace{5pt}%
	\begin{subfigure}[c]{0.07\columnwidth}
	\caption{}\label{fig:results-off-road-tpv-images-ours}
	\end{subfigure}%
	\begin{minipage}[c]{0.93\columnwidth}
	\includegraphics[width=0.322\columnwidth]{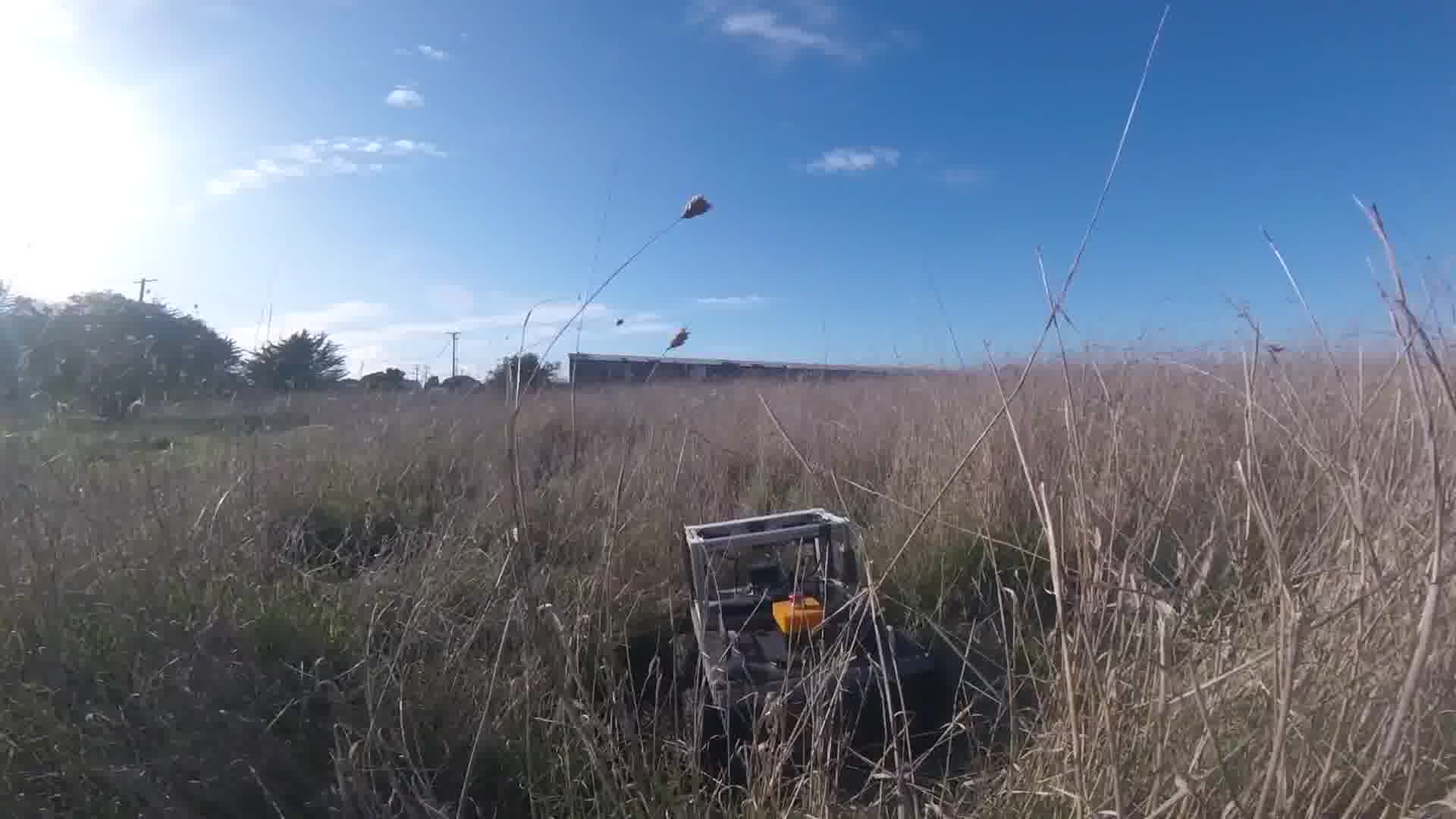}
	\hfill
	\includegraphics[width=0.322\columnwidth]{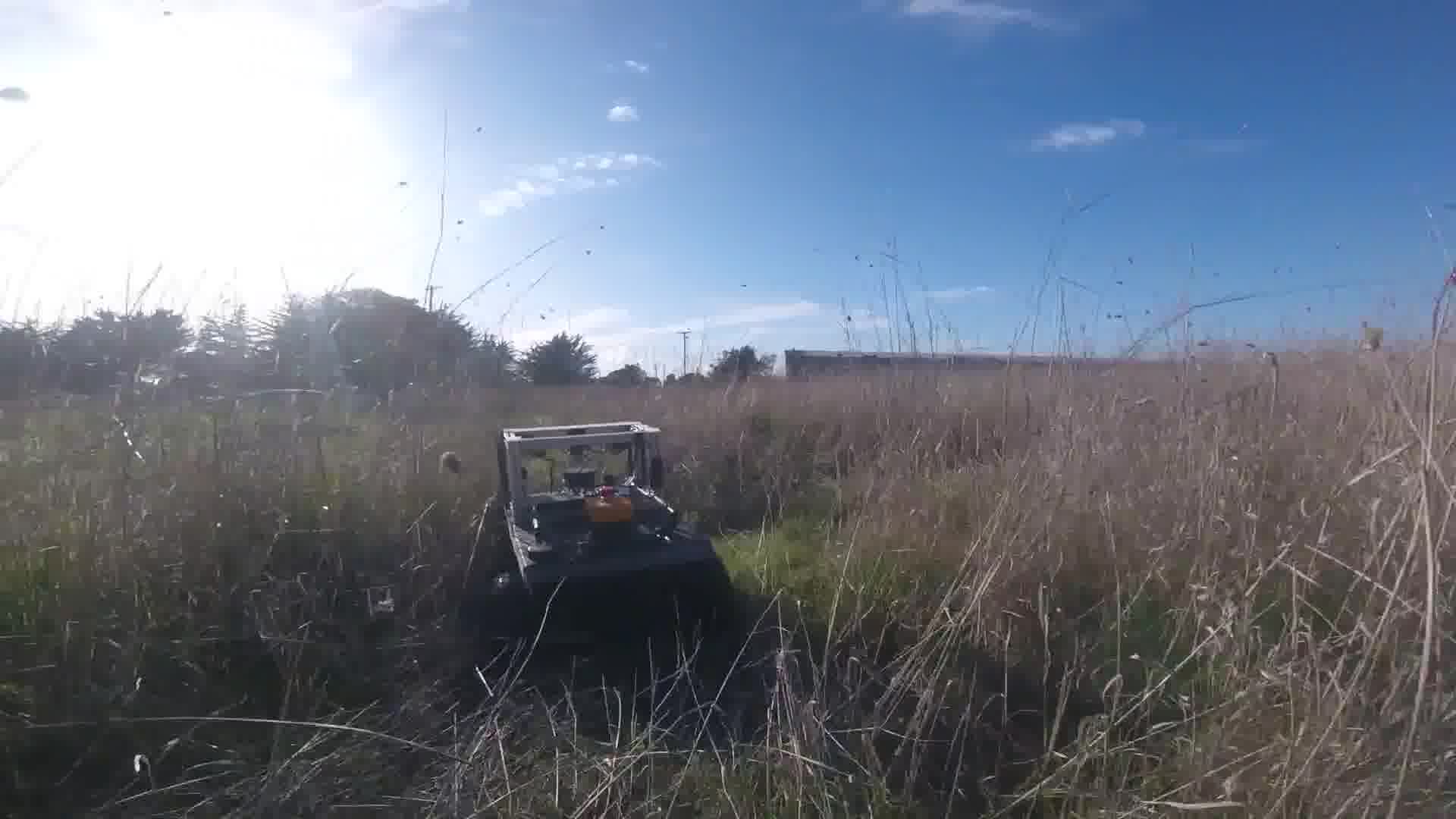}
	\hfill
	\includegraphics[width=0.322\columnwidth]{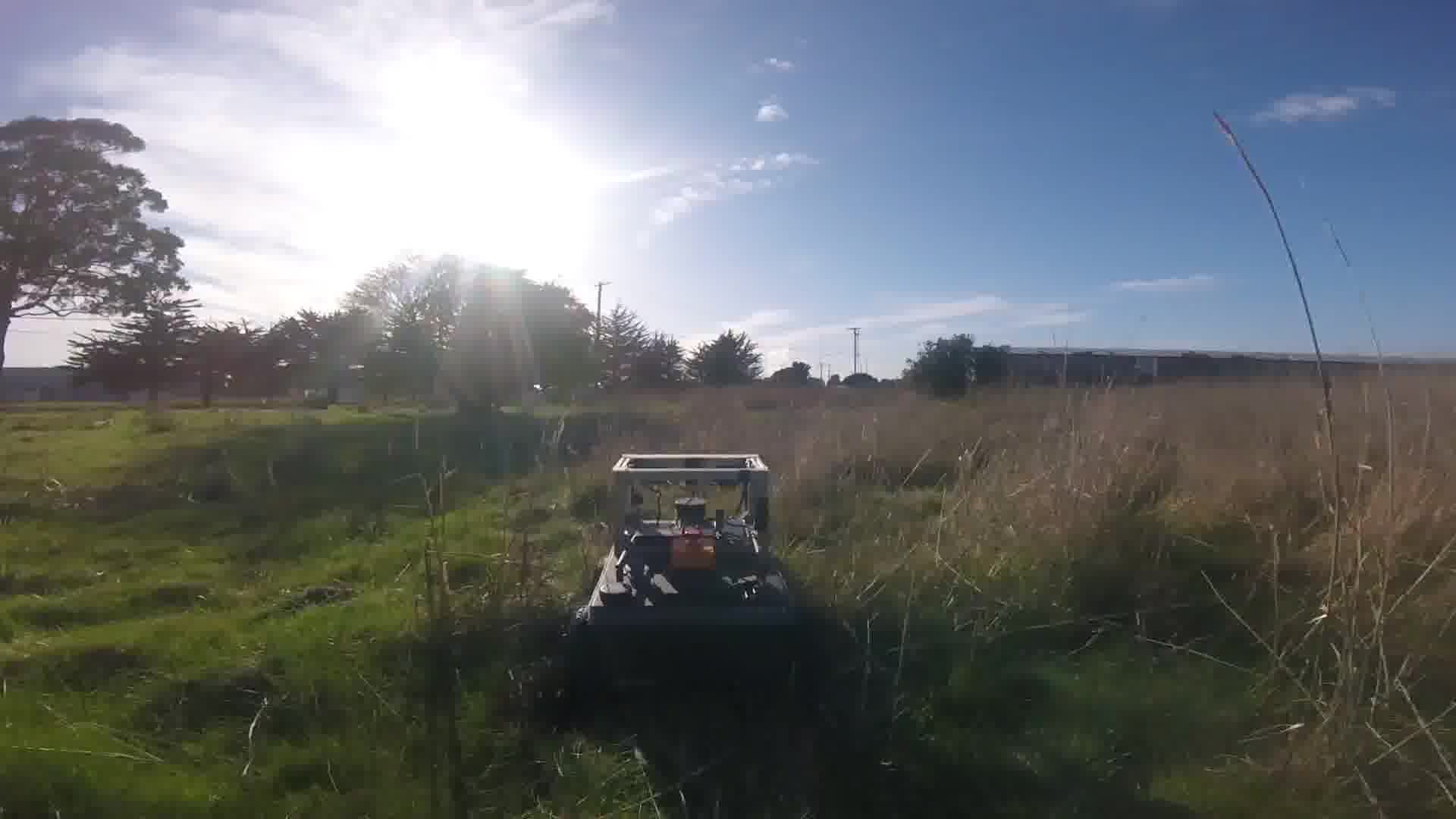}
	\end{minipage}
	\caption{Comparison of LIDAR (top) versus our \badgr approach (bottom) in a tall grass portion of the off-road environment. The LIDAR policy incorrectly labels the grass as untraversable, and therefore rotates in-place in an attempt to find a traversable path; after completing a full rotation and failing to detect any traversable path, the LIDAR policy determines the robot is trapped. In contrast, our \badgr approach has learned from experience that some tall grass is indeed traversable, and is therefore able to successfully navigate the robot towards the goal.}	
	\label{fig:results-off-road-tpv-images}
	\vspace*{-15pt}
\end{figure}

\textbf{Off-road environment.}
Next, we evaluated all the approaches for the task of reaching a goal GPS location while avoiding both collisions and getting stuck in an off-road environment. Fig.~\ref{fig:results-off-road-paths} shows the resulting paths that \badgr, LIDAR, and the na\"ive policies followed. The na\"ive policy sometimes succeeded, but oftentimes collided with obstacles such as trees and became stuck on thick patches of grass. The LIDAR policy nearly never crashed or became stuck on grass, but sometimes refused to move because it was surrounded by grass which it incorrectly labelled as untraversable obstacles (Fig.~\ref{fig:results-off-road-tpv-images-lidar}). \badgr almost always succeeded in reaching the goal by avoiding collisions and getting stuck, while not falsely predicting that all grass was an obstacle (Fig.~\ref{fig:results-off-road-tpv-images-ours}).

\begin{figure}[t]
    \centering
	\includegraphics[height=0.165\textheight,trim={27cm 2.0cm 45.3cm 2cm},clip]{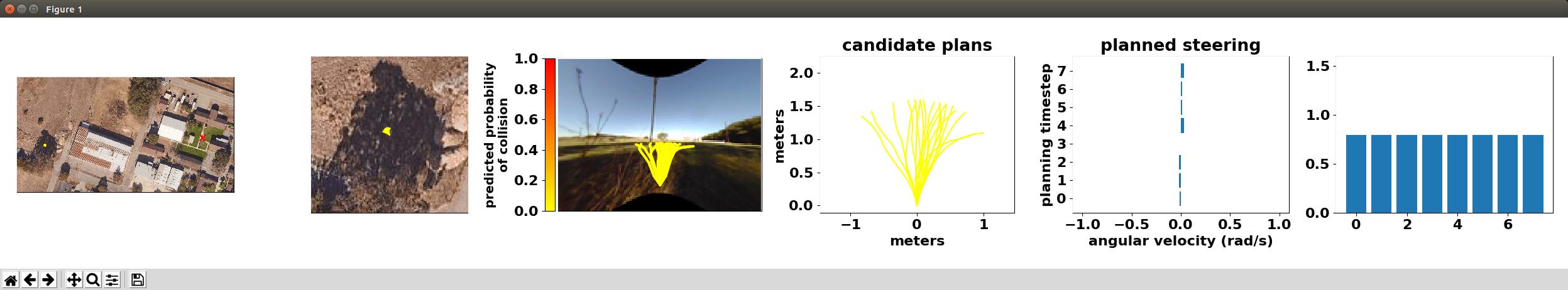}
	\hfill
	\includegraphics[height=0.165\textheight,trim={31.4cm 2.0cm 45.3cm 2cm},clip]{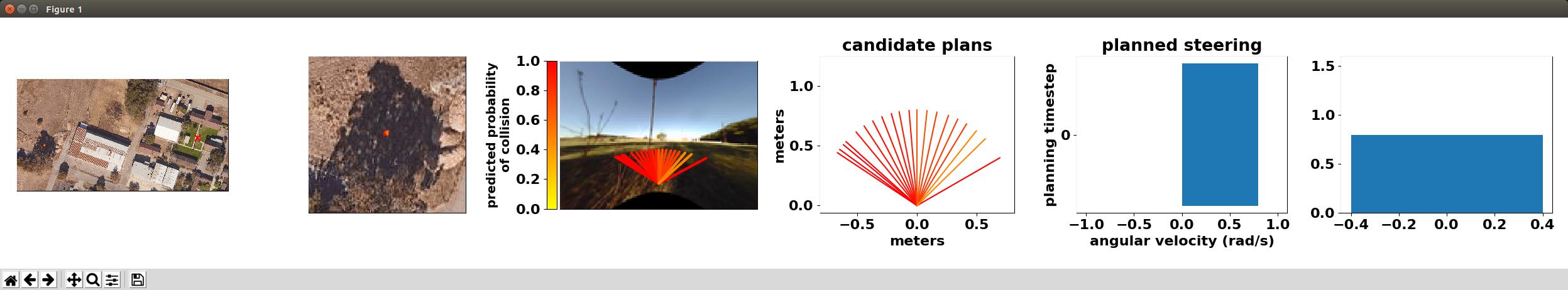}
    \caption{Comparison of our \badgr policy (left) versus the LIDAR policy (right). Each image shows the candidate paths each policy considered during planning, and the color of each path indicates if the policy predicts the path will result in a collision. The LIDAR policy falsely predicts the paths driving left or straight will result in a collision with the few strands of tall grass. In contrast, our \badgr policy correctly predicts that the grass is traversable and will therefore drive over the grass, which will result in \badgr reaching the goal $1.5 \times$ faster.}
    \label{fig:results-actionselection-off-road-vs-lidar}
    \vspace*{-15pt}
\end{figure}

Additionally, even when the LIDAR approach succeeded in reaching the goal, the path it took was sometimes suboptimal. Fig.~\ref{fig:results-actionselection-off-road-vs-lidar} shows an example where the LIDAR policy labelled a few strands of grass as untraversable obstacles, and therefore decided to take a roundabout path to the goal; in contrast, \badgr accurately predicted these few strands of grass were traversable, and therefore took a more optimal path. \badgr reached the goal $1.2 \times$ faster on average compared to the LIDAR policy.

\textbf{Self-improvement.}
A practical deployment of \badgr would be able to continually self-supervise and improve the model as the robot gathers more data. To provide an initial evaluation of how additional data enables adaptation to new circumstances, we conducted a controlled study in which \badgr gathers and trains on data from one area, moves to a new target area, fails at navigating in this area, but then eventually succeeds in the target area after gathering and training on additional data from that area. 

\begin{figure}[b]
    \vspace*{-10pt}
    \centering
    \includegraphics[width=\columnwidth]{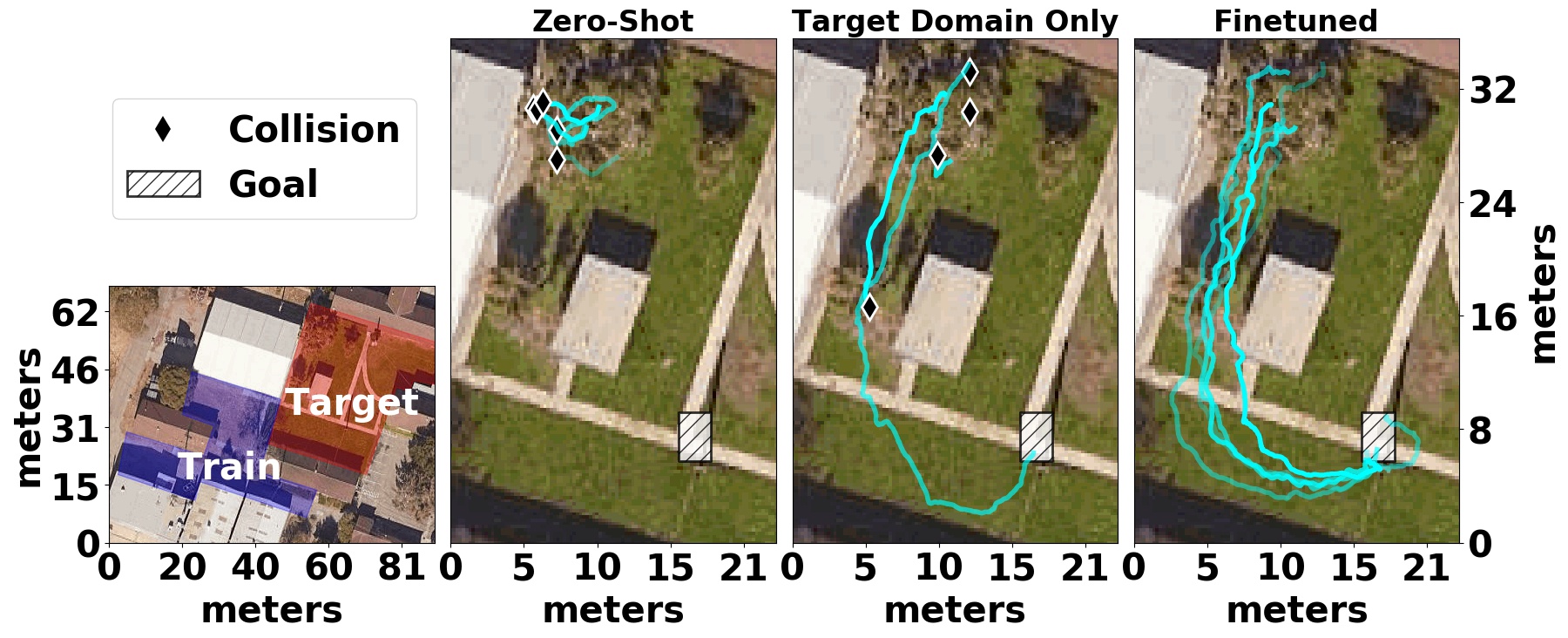}
    \caption{Experimental demonstration of our \badgr approach improving as it gathers more experience.     The robot's task was to reach a goal in the new target area without colliding. The zero-shot policy trained with only the initial training data always failed. The finetuned policy trained only using data from the target domain travelled farther before colliding, but still predominantly failed. The policy trained using both the initial training data and three hours of autonomously gathered, self-supervised-supervised data in the target domain always succeeded in reaching the goal. This result demonstrates that \badgr improves as it gathers more data, and also that previously gathered data from other areas can actually accelerate learning.}   
    \label{fig:results-selfimproving-paths}
\end{figure}

In this experiment, we first evaluate the performance of the original model trained only in the initial training domain, labeled as `zero-shot' in Figure~\ref{fig:results-selfimproving-paths}. The zero-shot policy fails on every trial due to a collision. We then evaluate the performance of a policy that is finetuned after collecting three more hours of data with autonomous self-supervision in the target domain, which we label as `finetuned.' This model succeeds at reaching the goal on every trial. For completeness, we also evaluate a model trained \emph{only} on the data from the target domain, without using the data from the original training domain, which we label as `target domain only.' This model is better than the zero-shot model, but still fails much more frequently than the finetuned model that uses both sources of experience.

This experiment not only demonstrates that \badgr can improve as it gathers more data, but also that previously gathered experience can actually accelerate policy learning when \badgr encounters a new environment. From these results, we might reasonably extrapolate that as \badgr gathers data in more and more environments, it should take less and less time to successfully learn to navigate in each new environment; we hope that future work will evaluate these truly continual and lifelong learning capabilities.

\begin{figure*}[t]
    \centering
    \captionsetup[subfigure]{aboveskip=1pt,belowskip=-7pt}
    \includegraphics[width=1.0\textwidth]{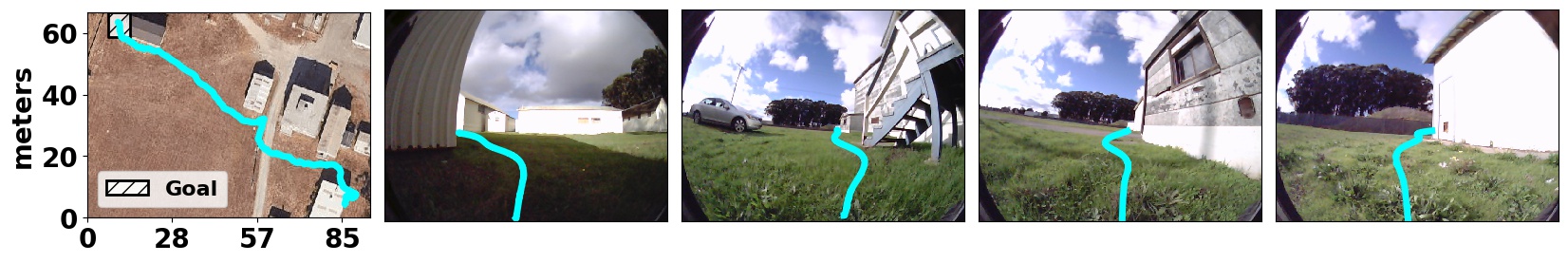}
	\includegraphics[width=1.0\textwidth]{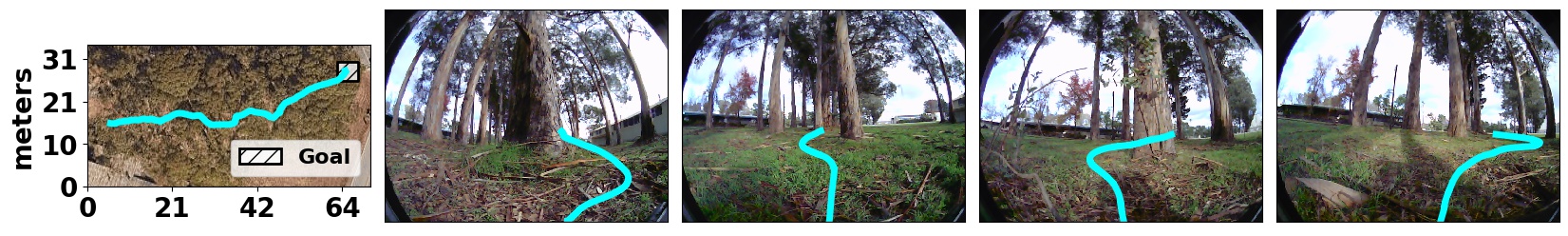}
	\includegraphics[width=1.0\textwidth]{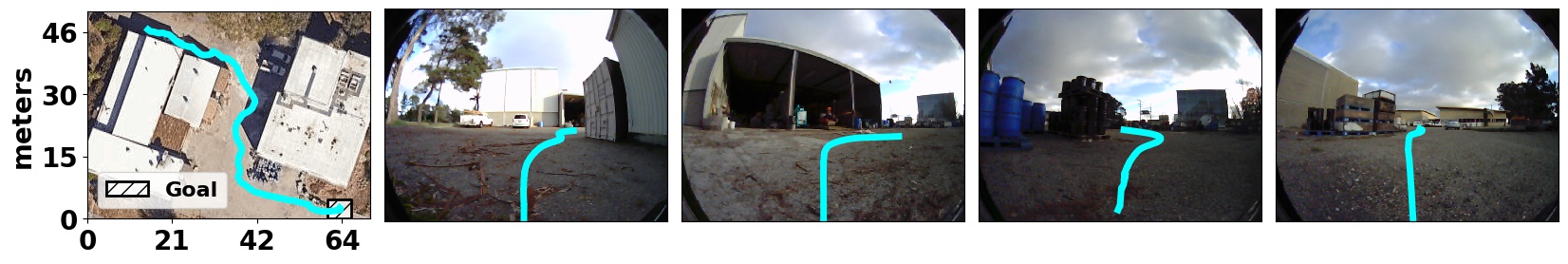}
    \caption{Our \badgr policy can generalize to novel environments not seen in the training data. Each row shows the \badgr policy executing in a different novel environment. The first column shows the approximate path followed by the \badgr policy. The remaining columns show sampled images from the onboard camera while the robot is navigating, with the future path of the robot overlaid onto the image.}
    \label{fig:results-generalization}
    \vspace*{-10pt}
\end{figure*}

\textbf{Generalization.}
We also evaluated how well {\badgr}---when trained on the full 42 hours of collected data---navigates in novel environments not seen in the training data. Fig.~\ref{fig:results-generalization} shows our \badgr policy successfully navigating in three novel environments, ranging from a forest to urban buildings. This result demonstrates that \badgr can generalize to novel environments if it gathers and trains on a sufficiently large and diverse dataset.

\section{Discussion} 
\label{sec:discussion}

We presented \badgr, an end-to-end learning-based mobile robot navigation system that can be trained entirely with self-supervised, off-policy data gathered in real-world environments, without any simulation or human supervision, and can improve as it gathers more data. We demonstrated that our approach can learn to navigate in real-world environments with geometrically distracting obstacles, such as tall grass, and can readily incorporate terrain preferences, such as avoiding bumpy terrain, using only 42 hours of data. Our experiments showed that \badgr can outperform a LIDAR policy in complex real-world settings, generalize to novel environments, and can improve as it gathers more data.

Although \badgr can autonomously gather data with minimal human supervision, the robot periodically requires human assistance, for example if the robot flips over. These periodic resets can be a significant burden, especially if the robot is operating in a remote location. Investigating methods that train policies which are cautious in novel environments could further decrease the amount of human supervision needed while collecting data. Also, while \badgr can improve as it gathers more data, this improvement requires gathering a non-negligible amount of data and retraining from scratch. Approaches that adapt in an online fashion could improve {\badgr}'s performance when deployed. Finally, our experiments solely evaluated \badgr in static environments, without moving agents such as pedestrians and cars. Gathering and training with data from non-static environments could prove challenging due to biases in the data from the interactions between the data collection policy and the other agents in the environment. We believe that solving these and other challenges is crucial for enabling robot learning platforms to learn and act in the real world, and that \badgr is a promising step towards this goal.


\section{Acknowledgments}

This work was supported by ARL DCIST CRA W911NF-17-2-0181, the National Science Foundation via IIS-1700697, the DARPA Assured Autonomy program, and Berkeley Deep Drive. Gregory Kahn was supported by an NSF Graduate Research Fellowship.


\bibliographystyle{unsrtnat}
\bibliography{references}

\end{document}